\documentclass[conference]{IEEEtran}
\IEEEoverridecommandlockouts
\usepackage{cite}
\usepackage{amsmath,amssymb,amsfonts}
\usepackage{algorithmic}
\usepackage{graphicx}
\usepackage{textcomp}
\usepackage{multirow}
\usepackage{booktabs}
\usepackage{subfigure}
\usepackage{xcolor}
\usepackage{threeparttable}
\usepackage{footnote}
\usepackage[hidelinks]{hyperref}
\usepackage[a4paper, total={184mm,239mm}]{geometry}

\usepackage{float}
\newcommand{\LLS}[1]{\textcolor{black}{#1}}

\def\BibTeX{{\rm B\kern-.05em{\sc i\kern-.025em b}\kern-.08em
    T\kern-.1667em\lower.7ex\hbox{E}\kern-.125emX}}
\begin{document}

\title{RLPlanner: Reinforcement Learning based Floorplanning for Chiplets with Fast Thermal Analysis
\thanks{This work was supported by Pre-research project of ministry foundation (Grant No.31513010501).(\textit{* Corresponding Authors: Leilai Shao (leilaishao@sjtu.edu.cn) and Xiaolei Zhu (xl\_zhu@zju.edu.cn)})}
}
\author{
\IEEEauthorblockN{Yuanyuan Duan$^{1}$, Xingchen Liu$^{1}$, Zhiping Yu$^{2}$, Hanming Wu$^{1}$, Leilai Shao$^{3*}$, Xiaolei Zhu$^{1*}$}
    \IEEEauthorblockA{$^{1}$School of Micro-Nano Electronics, Zhejiang University, Hangzhou, China}
    \IEEEauthorblockA{$^{2}$School of Integrated Circuits, Tsinghua University, Beijing, China}
    \IEEEauthorblockA{$^{3}$School of Mechanical Engineering, Shanghai Jiao Tong University, Shanghai, China}
    }

\maketitle

\begin{abstract}
Chiplet-based systems have gained significant attention in recent years due to their low cost and competitive performance. As the complexity and compactness of a chiplet-based system increase, careful consideration must be given to microbump assignments, interconnect delays, and thermal limitations during the floorplanning stage. This paper introduces \textit{RLPlanner}, an efficient \LLS{early-stage} floorplanning tool for chiplet-based systems with a novel fast thermal evaluation method. \textit{RLPlanner} employs advanced reinforcement learning to jointly minimize total wirelength and temperature. \LLS{To alleviate the time-consuming thermal calculations, \textit{RLPlanner} incorporates the developed fast thermal evaluation method to expedite the iterations and optimizations.} Comprehensive experiments demonstrate that our proposed \LLS{fast thermal evaluation method} achieves a mean absolute error (MAE) of ±0.25 K and delivers over 120x speed-up compared to the open-source thermal solver \textit{HotSpot}. When integrated with our fast thermal evaluation method, \textit{RLPlanner} achieves an average improvement of 20.28\% in minimizing the target objective (a combination of wirelength and temperature), within a similar running time, \LLS{compared to the classic simulated annealing method with \textit{HotSpot}}.
\end{abstract}

\begin{IEEEkeywords}
reinforcement learning, fast thermal evaluation, chiplet floorplanning
\end{IEEEkeywords}

\section{Introduction}
To address the increasing cost of large Systems-on-Chip (SoCs) on advanced technology nodes, chiplet-based design or 2.5D integration has emerged as a solution. 
However, as the complexity and compactness of chiplet-based systems increase, it becomes critical to address issues such as microbump assignments, interconnect delays, and thermomechanical stress during the initial floorplanning stage. 

Traditional physical floorplanning of monolithic chips primarily focuses on reducing total wirelength and minimizing area \cite{chen2006modern}, which results in compact floorplans that may lead to potential thermal-induced failures. Recent floorplanning works have started considering the thermal aspect in chiplet-based systems  \cite{coskun2018cross}. They employ simple greedy strategies or simulated annealing (SA) to handle chiplet system floorplanning, lacking the flexibility and transferability required for complex 2.5D integrations. Moreover, these optimization methods are constrained by time-consuming thermal evaluations, which hinder fast and efficient exploration of thermal-aware floorplanning. Some attempts have been made to use convolutional neural networks (CNN) or graph convolutional networks (GCN) to build surrogate models for accelerated thermal evaluations \cite{chen2022fast}. However, these models often rely on empirical parameters, such as tile and window sizes, requiring domain knowledge for determination, or show limited speed-ups ($<$3x), making them impractical for real designs.

\begin{figure}[t]
\centerline{\includegraphics[width=1\linewidth]{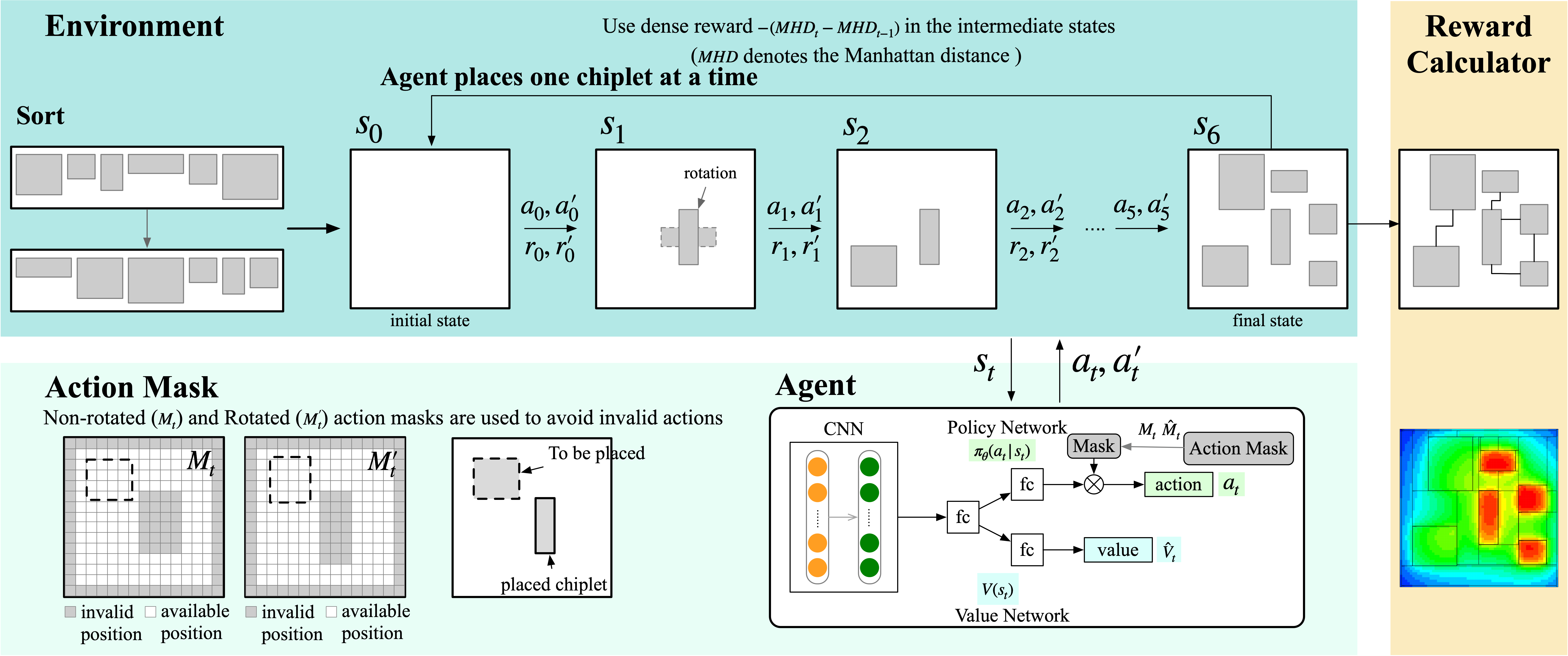}}
\caption{Overview of \textit{RLPlanner}: The implementation of chiplet floorplanning using reinforcement learning.}
\label{fig2}
\end{figure}

In this work, we present \textit{RLPlanner}, a floorplanning tool based on reinforcement learning (RL) for \LLS{early-stage} chiplet-based systems. \textit{RLPlanner} utilizes advanced RL techniques and incorporates a novel fast thermal evaluation method to optimize both the maximum operating temperature and total wirelength of the chiplet system. 

\begin{table*}[t]
\caption{Comparisons against baselines on benchmark systems}
\centering
\begin{threeparttable}
\tabcolsep=0.016\linewidth
\scalebox{0.69}{
\begin{tabular}{|l|c|c|c|c|c|c|c|c|c|c|c|c|}
\hline
\textbf{Design} & \multicolumn{4}{c}{\textbf{Multi-GPU System}\cite{ma2021tap}} & \multicolumn{4}{|c|}{\textbf{CPU-DRAM System}\cite{kannan2015enabling}} & \multicolumn{4}{c|}{\textbf{Ascend 910 System}\cite{Huawei910}} \\
\hline
\textbf{Method} & 
\textbf{Reward} & \begin{tabular}[c]{@{}c@{}}\textbf{Wirelength}\\ ($mm$)\end{tabular} & \begin{tabular}[c]{@{}c@{}}\textbf{Temperature}\\ ($^{\circ}$C)\end{tabular} & \begin{tabular}[c]{@{}c@{}}\textbf{Runtime}\\ ($s$)\end{tabular} & 
\textbf{Reward} & \begin{tabular}[c]{@{}c@{}}\textbf{Wirelength}\\ ($mm$)\end{tabular} & \begin{tabular}[c]{@{}c@{}}\textbf{Temperature}\\ ($^{\circ}$C)\end{tabular} & \begin{tabular}[c]{@{}c@{}}\textbf{Runtime}\\ ($s$)\end{tabular} & 
\textbf{Reward} & \begin{tabular}[c]{@{}c@{}}\textbf{Wirelength}\\ ($mm$)\end{tabular} & \begin{tabular}[c]{@{}c@{}}\textbf{Temperature}\\ ($^{\circ}$C)\end{tabular} & \begin{tabular}[c]{@{}c@{}}\textbf{Runtime}\\ ($s$)\end{tabular}  \\
\hline
\textit{RLPlanner} 		& -37.1263 & 97742 & 91.15 & 20910 & -44.9467 & 176246 & 92.88 &   8925 & -7.4063 & 18130 & 77.12 & 7773 \\ \hline
\textit{RLPlanner}(RND)   & -40.2777 & 104636  & 91.85 & 20380  & -41.7496 & 164460 & 92.15 & 8993 & -7.4433 & 18221 & 76.84 & 9991 \\ \hline
TAP-2.5D(\textit{HotSpot})      & -42.4572 & 124639 & 91.68 & 35211 & -60.3570 & 181269 & 97.94 & 15056 & -8.7651 & 21456 & 74.94 & 15731 \\  \hline
TAP-2.5D*(Fast Thermal Model) & -41.3358 & 111545 & 91.97 & 20782 & -50.2010 & 231859 & 92.82 & 9172 & -7.7890 & 19067 & 76.16 & 9984 \\ \hline

\end{tabular}
}
\begin{tablenotes}    
        \footnotesize               
        \item[*] takes a similar amount of time as training \textit{RLPlanner} for 600 epochs.  
      \end{tablenotes}            
    \end{threeparttable}       
\label{tab4}
\end{table*}

\section{Methodology}
\label{methodology}
\subsection{Overall Architecture of \textit{RLPlanner} }
\label{problem formulation}
As shown in Fig.~\ref{fig2}, the overall architecture of \textit{RLPlanner} consists of three main parts: the floorplanning environment for chiplets, the RL-based agent, and a thermal-aware reward calculator. The environment updates the action mask $M_{t}$ and state $s_t$ based on previously placed chiplets at each time step $t$. The agent generates the probability matrix of actions $\pi_{\theta}(a_t|s_t)$ and the expected reward $V(s_t)$ by the policy and value networks, correspondingly. Before sampling the action $a_t$ that places chiplets sequentially, the probability of infeasible actions will set to be '0' based on $M_{t}$. Once all chiplets have been placed, the reward calculator will first perform microbump assignments to minimize the total wirelength by allocating pin locations for each inter-chiplet connection, and then conduct the thermal evaluation.

\subsection{Reinforcement Learning and Agent Architecture}
In our agent architecture, the policy network and the value network share the same feature encoding CNN layers and two separate fully connected layers are used to get the probability matrix and expected reward. We employ the \LLS{proximal policy optimization} (PPO) \cite{schulman2017proximal} to train the networks. Moreover, we employ a random network distillation (RND) \cite{burda2018exploration} bonus to encourage the agent to explore novel states. It involves two neural networks: a fixed and randomly initialized target network, and a predictor network trained on data collected by the agent.

\subsection{Reward Calculations}
\label{Reward}
\begin{table}[t]
\caption{Accuracy and speed comparison during thermal evaluation}
\centering
\begin{threeparttable}
\scalebox{0.85}{
\begin{tabular}{|c|c|c|}
\hline
\textbf{Metrics} & \textbf{Fast Thermal Model} & \textbf{\textit{Hotspot}} \\ \hline
MSE  & 0.1732 K   & \multirow{ 4}{*}{\begin{tabular}[c]{@{}c@{}}Ground\\ Truth\end{tabular}} \\ \cline{1-2}
RMSE & 0.4162 K   &  \\ \cline{1-2}
MAE  & 0.2523 K   &  \\ \cline{1-2}
MAPE & 0.0726\% &  \\ \hline
\begin{tabular}[c]{@{}c@{}}Inference\\ Speed\end{tabular} & \begin{tabular}[c]{@{}c@{}}0.1012 s\\ (127$\times$)\end{tabular} &12.8976 s \\ \hline                 
\end{tabular}}
\end{threeparttable}
\label{tab3}
\end{table}

The target of chiplet floorplanning is to minimize the total wirelength and maximum operating temperature. We customize the reward function as $R=-\lambda \times W-\mu \times  \frac{(max(T-T_{0},0))^{\alpha}}{1+e^{-(T-T_{0})}}$ where $W$ and $T$ are the wirelength and temperature, $\lambda$ and $\mu$ are the weights, and $T_{0}$ is the temperature limitation and $\alpha$ is a hyperparameter to avoid gradient non-smoothness at $T=T_{0}$. The microbump assignments and wirelength optimization can be found in \cite{ma2021tap}. Traditional thermal analysis simulator \textit{HotSpot} \cite{huang2006hotspot} is CPU-intensive since it involves constructing a thermal model for the entire system and solving large linear equations repeatedly. To address the computational burden while maintaining accurate thermal analysis, a physical-informed fast thermal model is proposed. It simplifies the thermal resistance network structure by treating it as an linear and time-invariant (LTI) system. By calculating the self-thermal and mutual-thermal resistance in the thermal resistance network, the chiplets' temperatures can be obtained. Thus, we characterize the self-thermal resistance by setting a chiplet's power to a non-zero value and run \textit{HotSpot} to create a 2D self-thermal resistance table, and characterize the mutual-thermal resistance by a 1D table with respect to the distance between power source and grid location.  \LLS{More details can be found at the \href{https://github.com/weiweihook/RLPlanner}{github.}}


\section{Experimental Results}

\subsection{Quantitative evaluations of the fast thermal model}
A dataset comprising 2,000 synthetic chiplet systems are used to conduct thorough comparisons between the proposed fast thermal model and \textit{Hotspot}. The results are summarized in TABLE~\ref{tab3}, where mean square error (MSE), root mean square error (RMSE), mean absolute error (MAE) and mean absolute percentage error (MAPE) are calculated. The comparisons demonstrate that the proposed fast thermal model can accurately estimate the maximum temperature of a chiplet-based system and achieve a speedup over 127x during thermal evaluations compared to \textit{Hotspot}.

\begin{table}[]
\caption{Comparisons of reward on 5 synthetic systems}
\centering
\begin{threeparttable}
\tabcolsep=0.016\linewidth
\scalebox{0.85}{
\begin{tabular}{|l|c|c|c|c|c|c|}
\hline
 & \multicolumn{5}{c|}{\textbf{Reward}}   \\ \hline
\textbf{Method} & \textbf{Case1} & \textbf{Case2} & \textbf{Case3} & \textbf{Case4} & \textbf{Case5}   \\ \hline
\textit{RLPlanner}\  	   & -5.8288 & \textbf{-6.3236} & -10.0058 & -8.4076 & -8.6193\\ \hline
\textit{RLPlanner}(RND)& \textbf{-5.1062} & -6.7848& \textbf{-9.9335}& \textbf{-8.3903} & \textbf{-8.2049}  \\ \hline
TAP-2.5D(\textit{HotSpot})      			   & -6.6439 & -8.9846 & -12.3946 & -10.5525 & -10.6965\\ \hline
TAP-2.5D*(Fast Thermal Model) 				   & -6.3627 & -7.1250 & -10.7151 & -9.8286 & -8.5189 \\ \hline

\end{tabular}
}
\begin{tablenotes}    
        \footnotesize               
        \item[*] takes a similar amount of time as training \textit{RLPlanner} for 600 epochs. 
      \end{tablenotes}            
    \end{threeparttable}       

\label{tab5}
\end{table}

\subsection{Performance evaluations of \textit{RLPlanner}}
Three open-source benchmarks and five synthetic systems are used to evaluate the performance of \textit{RLPlanner}. 
Comparisons and analysis between the developed \textit{RLPlanner} and TAP-2.5D \cite{ma2021tap}, an SA-based thermal-aware floorplanning algorithm, are presented in TABLE~\ref{tab4} and TABLE~\ref{tab5}. Across all eight cases, \textit{RLPlanner} with RND achieves an average improvement of 20.28\% in optimization goals compared to TAP-2.5D with \textit{HotSpot}, and an average improvement of 9.25\% compared to TAP-2.5D with the fast thermal model within similar or less running times. It confirms our intuition that by conducting end-to-end co-optimizations, \textit{RLPlanner} shows better efficiency and quality in optimizing chiplet-based systems.


\bibliographystyle{IEEEtran}
\bibliography{DATE.bib}

\begin{thebibliography}{1}
\providecommand{\url}[1]{#1}
\csname url@samestyle\endcsname
\providecommand{\newblock}{\relax}
\providecommand{\bibinfo}[2]{#2}
\providecommand{\BIBentrySTDinterwordspacing}{\spaceskip=0pt\relax}
\providecommand{\BIBentryALTinterwordstretchfactor}{4}
\providecommand{\BIBentryALTinterwordspacing}{\spaceskip=\fontdimen2\font plus
\BIBentryALTinterwordstretchfactor\fontdimen3\font minus
  \fontdimen4\font\relax}
\providecommand{\BIBforeignlanguage}[2]{{%
\expandafter\ifx\csname l@#1\endcsname\relax
\typeout{** WARNING: IEEEtran.bst: No hyphenation pattern has been}%
\typeout{** loaded for the language `#1'. Using the pattern for}%
\typeout{** the default language instead.}%
\else
\language=\csname l@#1\endcsname
\fi
#2}}
\providecommand{\BIBdecl}{\relax}
\BIBdecl

\bibitem{chen2006modern}
T.-C. Chen \emph{et~al.}, ``Modern floorplanning based on b/sup*/-tree and fast
  simulated annealing,'' \emph{IEEE Transactions on Computer-Aided Design of
  Integrated Circuits and Systems}, vol.~25, no.~4, pp. 637--650, 2006.

\bibitem{coskun2018cross}
A.~Coskun \emph{et~al.}, ``A cross-layer methodology for design and
  optimization of networks in 2.5 d systems,'' in \emph{2018 IEEE/ACM
  International Conference on Computer-Aided Design (ICCAD)}.\hskip 1em plus
  0.5em minus 0.4em\relax IEEE, 2018, pp. 1--8.

\bibitem{chen2022fast}
L.~Chen \emph{et~al.}, ``Fast thermal analysis for chiplet design based on
  graph convolution networks,'' in \emph{2022 27th Asia and South Pacific
  Design Automation Conference (ASP-DAC)}.\hskip 1em plus 0.5em minus
  0.4em\relax IEEE, 2022, pp. 485--492.

\bibitem{ma2021tap}
Y.~Ma \emph{et~al.}, ``Tap-2.5 d: A thermally-aware chiplet placement
  methodology for 2.5 d systems,'' in \emph{2021 Design, Automation \& Test in
  Europe Conference \& Exhibition (DATE)}.\hskip 1em plus 0.5em minus
  0.4em\relax IEEE, 2021, pp. 1246--1251.

\bibitem{kannan2015enabling}
A.~Kannan \emph{et~al.}, ``Enabling interposer-based disintegration of
  multi-core processors,'' in \emph{Proceedings of the 48th international
  symposium on Microarchitecture}, 2015, pp. 546--558.

\bibitem{Huawei910}
\BIBentryALTinterwordspacing
{Huawei ascend 910 provides an nVidia AI training alternative}. [Online].
  Available:
  \url{https://www.servethehome.com/huawei-ascend-910-provides-a-nvidia-aitraining-
  alternative/}
\BIBentrySTDinterwordspacing

\bibitem{schulman2017proximal}
J.~Schulman \emph{et~al.}, ``Proximal policy optimization algorithms,''
  \emph{arXiv preprint arXiv:1707.06347}, 2017.

\bibitem{burda2018exploration}
Y.~Burda \emph{et~al.}, ``Exploration by random network distillation,''
  \emph{arXiv preprint arXiv:1810.12894}, 2018.

\bibitem{huang2006hotspot}
W.~Huang \emph{et~al.}, ``Hotspot: A compact thermal modeling methodology for
  early-stage vlsi design,'' \emph{IEEE Transactions on very large scale
  integration (VLSI) systems}, vol.~14, no.~5, pp. 501--513, 2006.

\end{thebibliography}

\end{document}